\documentclass[runningheads]{llncs}
\usepackage[T1]{fontenc}
\usepackage{graphicx}
\usepackage{amsmath}
\usepackage{amssymb}
\usepackage{booktabs}
\usepackage{graphicx}
\usepackage{multirow}
\usepackage{hyperref}
\usepackage{enumitem}
\usepackage{booktabs}
\usepackage{makecell}
\begin{document}
%
\title{Benchmarking MRI Representations for Deep Learning-Based Focal Cortical Dysplasia Segmentation}
%
%
\author{Soumen Ghosh\inst{1,2*}\and
John Phamnguyen\inst{1,6,7} \and
Amit Soni Arya \inst{3} \and
Subhojit Mandal\inst{4}\and
Tilottama Goswami\inst{5} \and
Rajat Vashistha \inst{1}}

%
\authorrunning{S. Ghosh et al.}
%
\titlerunning{Benchmarking MRI Representations for DL-Based FCD Segmentation}
\institute{The University of Queensland, Brisbane, Australia \and
I-MED Radiology, Brisbane, Australia \and
Bennett University, Greater Noida, India \and
Indian Institute of Technology Madras, India \and
University College of Engineering, Osmania University, Hyderabad, India \and
Mater Hospital, Brisbane, Australia \and
Royal Brisbane and Women’s Hospital, Brisbane, Australia \\
\email{*Corresponding author: soumen.ghosh@uq.edu.au}}
\maketitle              

\begin{abstract}
Focal cortical dysplasia (FCD) is one of the leading structural causes of drug-resistant focal epilepsy, yet its subtle and heterogeneous imaging characteristics make accurate identification and delineation challenging on conventional magnetic resonance imaging (MRI). Although T1-weighted (T1w) and fluid-attenuated inversion recovery (FLAIR) images are routinely acquired for presurgical evaluation, the contribution of different MRI representations to deep learning-based FCD segmentation remains poorly understood. In this study, we present a systematic benchmark of MRI representations for automated FCD segmentation using the nnU-Net framework. A publicly available presurgical MRI dataset comprising 85 FCD subjects and 25 healthy controls was used to evaluate eight input configurations, including conventional MRI contrasts (T1w and FLAIR), ratio-derived representations (T1w/FLAIR and FLAIR/T1w), and their multimodal combinations. To isolate the effect of MRI representation, all experiments employed identical preprocessing, network architecture, optimization strategy, and five-fold cross-validation. Among the evaluated single-modality representations, FLAIR achieved the strongest overall performance, whereas ratio-derived representations alone were insufficient for reliable identification of subtle FCD. Incorporating ratio-derived representations with conventional T1w and FLAIR images consistently improved lesion delineation, with the four-channel multimodal configuration achieving the highest overall Dice score (0.376), representing a 5.0\% relative improvement over the conventional T1w+FLAIR representation. These findings demonstrate that MRI representation design is an important yet underexplored component of deep learning-based FCD segmentation and should be optimized alongside network architecture. The proposed benchmark provides practical guidance for designing multimodal MRI inputs for future FCD segmentation methods.


\end{abstract}

\section{Introduction}
Epilepsy affects more than 65 million people worldwide, with focal cortical dysplasia (FCD) representing one of the leading structural causes of drug-resistant focal epilepsy \cite{ngugi2010estimation,taylor1971focal,hsieh2016convulsive}. For patients with medically refractory epilepsy, surgical resection of the epileptogenic lesion offers the best opportunity for seizure freedom when the epileptogenic zone can be accurately localized \cite{ryvlin2014epilepsy}. Consequently, accurate identification and delineation of FCD in the brain imaging are essential in presurgical planning \cite{kim2011neuroimaging,harvey2015surgically}.

Magnetic resonance imaging (MRI) is the primary imaging modality for detecting FCD \cite{bernasconi2001texture,spitzer2022interpretable}. T1-weighted (T1w) and fluid-attenuated inversion recovery (FLAIR) images provide complementary information regarding lesion morphology and tissue abnormalities \cite{gill2017automated}. T1w images highlight cortical thickening, abnormal gyration, and blurring of the gray-white matter junction, whereas FLAIR is more sensitive to lesion-associated signal abnormalities, including cortical-subcortical hyperintensity and the transmantle sign commonly observed in FCD Type II lesions \cite{sisodiya2009focal}. Despite advances in MRI acquisition and epilepsy-specific imaging protocols, subtle FCD remain challenging to detect, with approximately 30\% being overlooked on conventional MRI interpretation and many patients initially classified as MRI-negative, contributing to substantial inter-rater variability and delayed diagnosis \cite{urbach2022mri}.
These challenges have motivated the development of automated image analysis techniques for FCD detection and segmentation\cite{hong2014automated,snell2026mri}.

Recent advances in deep learning have substantially improved automated medical image segmentation \cite{ronneberger2015u,isensee2021nnu}. In epilepsy imaging, convolutional neural networks and U-Net-based architectures have demonstrated promising performance for FCD detection and segmentation using structural MRI \cite{gill2021multicenter,joshi2025nnu,ding2025automated,thomas2020multi}. More recently, nnU-Net has emerged as a strong benchmark framework owing to its robust performance and automatic adaptation to diverse medical imaging datasets \cite{isensee2021nnu}. However, most previous studies have focused on developing increasingly sophisticated network architectures and optimization strategies, while largely treating MRI representation design as a fixed preprocessing step. Consequently, the influence of MRI representation itself on deep learning-based FCD segmentation remains poorly understood.

Beyond conventional structural MRI, ratio-derived representations such as T1w/FLAIR have been proposed to enhance tissue contrast, cortical myelination patterns, and gray-white matter differentiation \cite{cappelle2022t1w,colaes2026evaluating}. Previous studies have demonstrated their utility for visualizing subtle structural abnormalities in epilepsy, suggesting that they may provide complementary information beyond conventional MRI sequences. However, their contribution to deep learning-based FCD segmentation has not been systematically investigated. In particular, it remains unclear whether ratio-derived representations are effective as standalone inputs or whether they provide additional benefit when integrated with conventional T1w and FLAIR images.

From a machine learning perspective, multimodal FCD segmentation depends not only on network architecture but also on how complementary MRI information is represented at the network input.
Rather than proposing another segmentation architecture, this study investigates whether optimizing MRI representation alone can improve FCD segmentation performance. By systematically evaluating conventional and ratio-derived MRI representations under an identical nnU-Net framework, we isolate the contribution of MRI representation design independently of architectural variation. The main contributions of this work are as follows:

\begin{enumerate}
\item We present the first systematic benchmark of conventional and ratio-derived MRI representations for deep learning-based FCD segmentation under a controlled experimental framework.
\item We investigate the complementary role of asymmetric ratio-derived MRI representations and demonstrate their contribution beyond conventional T1w and FLAIR imaging.
\item We demonstrate that MRI representation design is a complementary direction to network architecture development for improving automated FCD segmentation.
\end{enumerate}

\begin{figure*}[t]
    \centering
    \includegraphics[width=\textwidth]{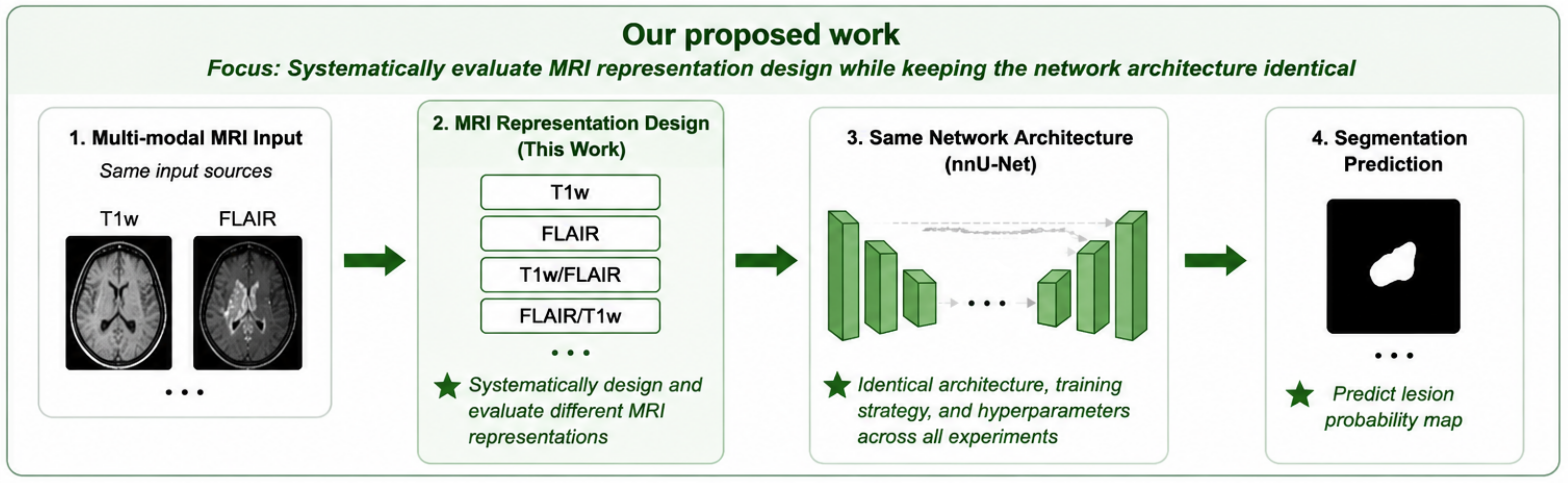}
    \caption{
    Conceptual overview of the proposed study.
    }
    \label{fig:model}
\end{figure*}

Figure \ref{fig:model} conceptually illustrates the motivation of this work. Unlike previous FCD segmentation studies, which primarily focus on developing increasingly sophisticated network architectures while treating MRI representation as a fixed preprocessing step, this study systematically investigates MRI representation design under an identical segmentation framework. A controlled experimental design was adopted in which the nnU-Net architecture, preprocessing, optimization strategy, and cross-validation protocol were kept identical across all experiments, allowing the contribution of MRI representations to be evaluated independently.

\section{Methodology}
\subsection{Dataset}

This study used the publicly available FCD dataset introduced by Schuch et al. \cite{schuch2023open}, which contains MRI scans from 85 subjects with focal cortical dysplasia (FCD) and 85 neurologically normal healthy controls (HC). The FCD cohort comprised 35 females and 50 males with a mean age of $28.9 \pm 12.4$ years, while the healthy controls had a mean age of $33.3 \pm 11.9$ years. Since lesion segmentation is performed only for subjects with FCD, all 85 FCD subjects were included in this study. To expose the model to normal anatomical variability and evaluate false-positive predictions, a randomly selected subset of 25 healthy controls was additionally included during training and evaluation. Expert-manually delineated lesion masks were available for all FCD subjects, whereas healthy controls were assigned empty lesion masks. For each subject, structural T1w and FLAIR MRI scans were used.


\subsection{Image Preprocessing}
\begin{figure*}[t]
    \centering
    \includegraphics[width=\textwidth]{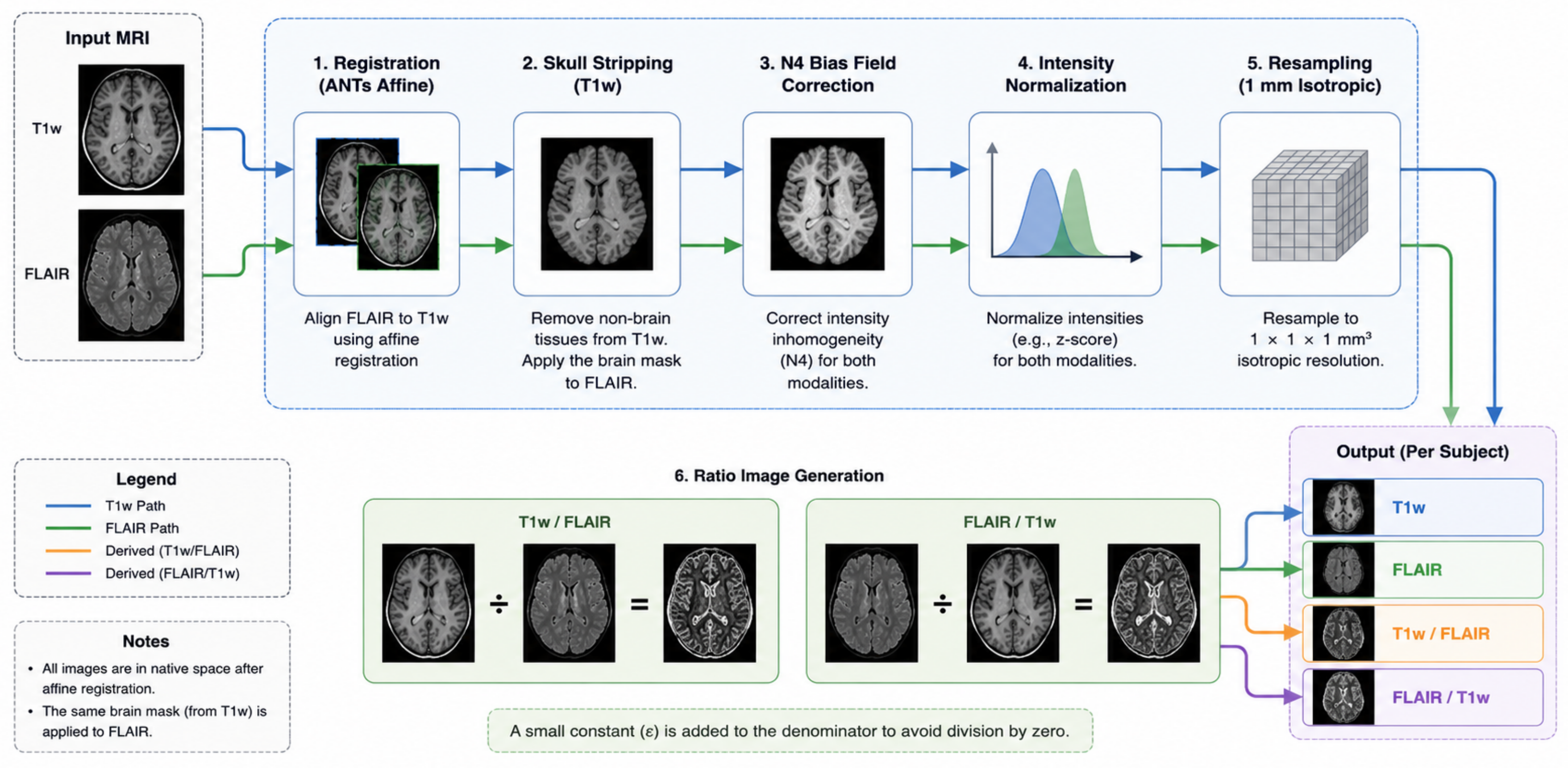}
    \caption{
    Overview of the proposed MRI preprocessing pipeline.
    }
    \label{fig:preprocessing}
\end{figure*}

All MRI volumes underwent a standardized preprocessing pipeline prior to model training (Figure \ref{fig:preprocessing}). First, T1-weighted (T1w) and FLAIR images were spatially aligned using affine registration implemented in the Advanced Normalization Tools (ANTs) framework to ensure voxel-wise correspondence between modalities. Following registration, skull stripping was performed on the T1w image to remove non-brain tissues. The resulting brain mask was subsequently applied to both T1w and FLAIR images to maintain consistent brain extraction across modalities.

To reduce intensity inhomogeneity arising from magnetic field non-uniformities, N4 bias field correction was applied independently to the skull-stripped T1w and FLAIR images. The corrected images were then resampled to an isotropic voxel resolution of $1 \times 1 \times 1 mm^3$ to ensure consistent spatial resolution across all subjects. Intensity normalization was subsequently performed on each modality to reduce inter-subject intensity variability and facilitate stable network training.

In addition to the conventional MRI contrasts, two ratio-derived representations were generated to investigate their contribution to FCD segmentation. Specifically, the T1w/FLAIR ratio image and the inverse FLAIR/T1w ratio image were computed on a voxel-wise basis using the normalized MRI volumes. These ratio representations were used either individually or in combination with the original MRI modalities according to the experimental configuration. The final preprocessed dataset therefore consisted of four image representations: T1w, FLAIR, T1w/FLAIR, and FLAIR/T1w, which were subsequently used as input channels for the nnU-Net models.


\subsection{Problem Formulation}
Given paired T1-weighted (T1w) and FLAIR MRI volumes, the objective is to predict a voxel-wise FCD mask. 

Let
\begin{equation}
X_T \in \mathbb{R}^{H \times W \times D}
\end{equation}
denote the T1w MRI volume and

\begin{equation}
X_F \in \mathbb{R}^{H \times W \times D}
\end{equation}
denote the FLAIR MRI volume, where (H), (W), and (D) represent the spatial dimensions of the MRI volume.

To investigate the contribution of MRI representations, two ratio-derived images (T1w/FLAIR and FLAIR/T1w) are additionally constructed and evaluated individually or in combination with the original MRI modalities.
The T1w/FLAIR ratio image is defined as
\begin{equation}
R_{TF}
=
\frac{X_T}{X_F + \epsilon}
\end{equation}
and the inverse ratio image is

\begin{equation}
R_{FT}
=
\frac{X_F}{X_T + \epsilon}
\end{equation}
where $\epsilon$ is a small constant introduced to avoid numerical instability.

Depending on the experimental configuration, the network input is represented as a multi-channel tensor
\begin{equation}
    X=[X_1,X_2,\ldots,X_C],
\end{equation}
where $(C)$ denotes the number of input channels and $(X_i)$ corresponds to one of the MRI representations $({X_T, X_F, R_{TF}, R_{FT}})$.

Let
\begin{equation}
    Y \in {0,1}^{H \times W \times D}
\end{equation}
denote the ground-truth lesion mask, where $(Y(v)=1)$ indicates lesion voxels and $(Y(v)=0)$ indicates background voxels. The objective is to learn a segmentation function
\begin{equation}
    f_{\theta}: X \rightarrow \hat{Y},
\end{equation}
parameterized by $(\theta)$, that maps the input MRI representation $(X)$ to a predicted lesion probability map
\begin{equation}
    \hat{Y}=f_{\theta}(X).
\end{equation}

The optimal model parameters are obtained by minimizing a supervised segmentation loss over the training dataset
\begin{equation}
    \arg\min_{\theta}
\frac{1}{N}
\sum_{n=1}^{N}
\mathcal{L}
\left(
f_{\theta}(X_n),
Y_n
\right),
\end{equation}
where $(N)$ denotes the number of training subjects and $(\mathcal{L})$ represents the segmentation loss function. In this study, the nnU-Net framework is employed to learn the mapping between MRI representations and FCD masks, while maintaining identical network architecture and optimization settings across all experiments.

\subsection{nnU-Net Framework}
To ensure a fair and standardized evaluation of MRI representations, all experiments were performed using the official nnU-Net v2 \cite{isensee2021nnu} framework with the 3D full-resolution configuration. nnU-Net is a widely adopted self-configuring medical image segmentation framework that automatically adapts the network architecture, preprocessing, data augmentation, and training parameters to the characteristics of a given dataset. It was selected as a strong baseline to enable a controlled comparison of MRI representation strategies while eliminating architectural confounding.

To ensure a fair comparison between MRI representations, all experiments employed identical network architecture, training schedules, and optimization settings. Models were trained for 1000 epochs using the five-fold cross-validation. The best-performing checkpoint based on validation performance was retained for subsequent evaluation.
Model optimization was performed using a compound objective function consisting of Dice loss and cross-entropy loss:

\begin{equation}
L= L_{\mathrm{Dice}}
+
\lambda L_{\mathrm{CE}},
\end{equation}

where

\begin{equation}
L_{\mathrm{Dice}}
=
1-
\frac{
2\sum_{i=1}^{V} p_i g_i + \delta
}{
\sum_{i=1}^{V} p_i + \sum_{i=1}^{V} g_i + \delta
},
\end{equation}

and

\begin{equation}
L_{\mathrm{CE}}
=
-
\sum_{i=1}^{V}
g_i \log(p_i).
\end{equation}

Here, \(p_i\) and \(g_i\) denote the predicted probability and ground-truth label for voxel \(i\), respectively, \(V\) is the total number of voxels, and \(\delta\) is a smoothing constant. The combined objective simultaneously optimizes lesion overlap and voxel-wise classification accuracy, which is advantageous for segmenting small FCD in the presence of substantial class imbalance between lesion and background voxels.

\section{Experimental Design}
\subsection{MRI Representation Experiments}
The primary objective of this study was to systematically investigate the contribution of conventional MRI contrasts and ratio-derived representations to automated FCD segmentation. To achieve this, eight MRI representation configurations (Table-\ref{tab:experiments}) were evaluated while maintaining identical preprocessing, network architecture, training strategy, and evaluation protocols across all experiments. The investigated configurations included single-modality inputs, ratio-derived representations, conventional multimodal inputs, and multimodal inputs augmented with ratio information.
\begin{table}[t]
\centering
\caption{MRI representation configurations and experimental objectives.}
\label{tab:experiments}
\footnotesize
\begin{tabular}{c|l|p{7.8cm}}
\hline
ID & Input & Objective \\
\hline
E1 & T1w &
Anatomical representation baseline. \\

E2 & FLAIR &
Lesion-sensitive representation baseline. \\

E3 & T1w/FLAIR &
Evaluate cortical contrast enhancement. \\

E4 & FLAIR/T1w &
Evaluate lesion-signal enhancement. \\

E5 & T1w + FLAIR &
Conventional multimodal baseline. \\

E6 & E5 + T1w/FLAIR &
Assess complementary cortical contrast information. \\

E7 & E5 + FLAIR/T1w &
Assess complementary lesion-signal information. \\

E8 & E6 + FLAIR/T1w &
Evaluate combined ratio representations. \\
\hline
\end{tabular}
\end{table}

The first four experiments evaluated the discriminative capability of individual MRI representations. Specifically, E1 and E2 assessed the performance of conventional T1w and FLAIR images, respectively, while E3 and E4 investigated the utility of ratio-derived representations obtained from T1w/FLAIR and FLAIR/T1w transformations. The remaining experiments explored multimodal fusion strategies. E5 served as the conventional multimodal baseline using T1w and FLAIR images, whereas E6 and E7 evaluated the incremental contribution of a single ratio-derived representation. Finally, E8 incorporated both ratio images in addition to the original MRI modalities, enabling assessment of whether complementary information from asymmetric ratio representations could further improve segmentation performance.

This experimental design enables analysis of three key research questions: (i) which individual MRI representation provides the strongest segmentation performance, (ii) whether ratio-derived representations offer complementary information beyond conventional MRI contrasts, and (iii) whether combining multiple ratio representations leads to further performance improvements over standard multimodal MRI inputs. Importantly, all experiments differ only in MRI representation, enabling the contribution of representation design to be evaluated independently of network architecture.

\subsection{Cross-Validation}

A five-fold stratified cross-validation strategy was employed, preserving the relative distribution of FCD and control subjects across folds. Each validation fold contained approximately 17 FCD subjects and 5 control subjects, with the remaining subjects used for training. To ensure a fair comparison, identical cross-validation splits were reused across all experiments (E1--E8), such that performance differences reflected only the MRI representation being evaluated. Model performance was averaged across the five folds and reported as mean \(\pm\) standard deviation. Control subjects were assigned empty lesion masks and included in both training and validation folds to assess model specificity and false-positive behaviour.

\subsection{Evaluation Metrics}
Segmentation performance was evaluated using the Dice Similarity Coefficient (Dice), sensitivity, and precision. Dice quantified the spatial overlap between the predicted and reference lesion masks, while sensitivity and precision measured lesion recovery and false-positive behaviour at the voxel level, respectively.

To provide a comprehensive assessment of model performance, two complementary analyses were performed. First, overall segmentation performance was evaluated across all FCD subjects, with undetected lesions assigned a Dice score of zero to reflect real-world detection performance. Second, segmentation accuracy was assessed only for successfully detected lesions to evaluate delineation performance independently of lesion detection. In addition, the numbers of completely missed FCD lesions (Dice = 0) and healthy control subjects with false-positive predictions were reported to quantify lesion detection failure and model specificity.

\subsection{Statistical Analysis}

Statistical analyses were performed to compare segmentation performance across the eight MRI representation configurations. Overall differences between experiments were assessed using the Friedman test, followed by pairwise Wilcoxon signed-rank tests with false discovery rate (FDR) correction for multiple comparisons.
In addition to statistical significance, effect sizes were quantified using Cohen's \(d\) to assess the magnitude of performance differences between MRI representations. Statistical significance was defined as \(p < 0.05\).

\section{Results}

\subsection{Overall Segmentation Performance}

Table~\ref{tab:overall_results} summarizes the segmentation performance of all MRI representation strategies across the complete evaluation cohort, including subjects for which no lesion was detected (Dice = 0).
Among the single-modality inputs, FLAIR achieved the highest Dice score (0.369$\pm$0.303), substantially outperforming T1w (0.246$\pm$0.290) and the T1w/FLAIR ratio representation (0.128$\pm$0.245). FLAIR also produced the lowest lesion miss rate (28\%), whereas T1w and T1w/FLAIR failed to detect 51\% and 73\% of lesions, respectively. These findings indicate that lesion-related signal abnormalities captured by FLAIR are considerably more informative for lesion localization than anatomical information alone.

\begin{table*}[t]
\centering
\footnotesize
\setlength{\tabcolsep}{3pt}
\renewcommand{\arraystretch}{1.05}
\caption{Overall segmentation performance across all subjects.}
\begin{tabular}{llcccccc}
\toprule
Exp & Input & Ch. & Dice & Sens. & Prec. &
\makecell{FCD Misses\\($\mathrm{Dice}=0$)} &
\makecell{Control\\FP} \\
\midrule
E1 & T1w                        & 1 & 0.246$\pm$0.290 & 0.235$\pm$0.302 & 0.365$\pm$0.397 & 0.51 & 0.40 \\
E2 & FLAIR                      & 1 & 0.369$\pm$0.303 & 0.352$\pm$0.316 & 0.535$\pm$0.396 & 0.28 & 0.20 \\
E3 & T1w/FLAIR                  & 1 & 0.128$\pm$0.245 & 0.124$\pm$0.250 & 0.184$\pm$0.336 & 0.73 & 0.08 \\
E4 & FLAIR/T1w                  & 1 & 0.333$\pm$0.319 & 0.323$\pm$0.334 & 0.438$\pm$0.403 & 0.40 & 0.20 \\
E5 & T1w+FLAIR                  & 2 & 0.358$\pm$0.314 & 0.341$\pm$0.324 & 0.512$\pm$0.406 & 0.33 & 0.12 \\
E6 & E5+T1w/FLAIR               & 3 & 0.367$\pm$0.315 & 0.345$\pm$0.315 & 0.490$\pm$0.408 & 0.35 & 0.20 \\
E7 & E5+FLAIR/T1w               & 3 & 0.361$\pm$0.308 & 0.346$\pm$0.326 & 0.492$\pm$0.403 & 0.33 & 0.16 \\
E8 & E5+T1w/FLAIR+FLAIR/T1w     & 4 & 0.376$\pm$0.323 & 0.356$\pm$0.322 & 0.496$\pm$0.406 & 0.34 & 0.20 \\
\bottomrule
\end{tabular}
\label{tab:overall_results}
\end{table*}

Among the multimodal representations, all configurations achieved comparable overall performance, with Dice scores ranging from 0.358 to 0.376. The full multimodal representation (E8) achieved the highest overall Dice score (0.376$\pm$0.323), suggesting that ratio-derived representations provide complementary information when combined with conventional MRI sequences. The conventional multimodal baseline (E5) yielded the lowest false-positive rate among controls (12\%), while E8 demonstrated the highest lesion sensitivity (0.356$\pm$0.322).

\subsection{Segmentation Performance for Detected Lesions}
To evaluate lesion delineation independently of lesion detection, segmentation accuracy was further assessed using only successfully detected lesions.
As shown in Table~\ref{tab:detected_results}, segmentation performance increased substantially for all MRI representations after excluding missed lesions. FLAIR/T1w achieved the highest Dice score among single-modality inputs (0.554$\pm$0.216), outperforming both T1w and FLAIR. Among multimodal representations, E8 achieved the highest Dice score (0.571$\pm$0.218), closely followed by E6 (0.567$\pm$0.199).

\begin{table}[t]
\centering
\footnotesize
\setlength{\tabcolsep}{4pt}
\renewcommand{\arraystretch}{1.05}
\caption{Performance comparison for detected FCD cases only (Dice $>$ 0).}
\label{tab:detected_results}
\begin{tabular}{llcccc}
\toprule
Exp & Input & Ch. & Dice & Sens. & Prec. \\
\midrule
E1 & T1w                    & 1 & 0.497$\pm$0.214 & 0.475$\pm$0.266 & 0.734$\pm$0.216 \\
E2 & FLAIR                  & 1 & 0.514$\pm$0.230 & 0.491$\pm$0.267 & 0.746$\pm$0.248 \\
E3 & T1w/FLAIR              & 1 & 0.473$\pm$0.241 & 0.457$\pm$0.280 & 0.682$\pm$0.281 \\
E4 & FLAIR/T1w              & 1 & 0.554$\pm$0.216 & 0.538$\pm$0.266 & 0.730$\pm$0.240 \\
E5 & T1w+FLAIR              & 2 & 0.534$\pm$0.232 & 0.508$\pm$0.268 & 0.764$\pm$0.232 \\
E6 & E5+T1w/FLAIR           & 3 & 0.567$\pm$0.199 & 0.534$\pm$0.229 & 0.757$\pm$0.234 \\
E7 & E5+FLAIR/T1w           & 3 & 0.538$\pm$0.215 & 0.515$\pm$0.266 & 0.734$\pm$0.255 \\
E8 & E5+T1w/FLAIR+FLAIR/T1w & 4 & 0.571$\pm$0.218 & 0.541$\pm$0.230 & 0.754$\pm$0.238 \\
\bottomrule
\end{tabular}
\end{table}

Compared with the conventional multimodal baseline (E5), incorporating ratio-derived representations consistently improved Dice and sensitivity, indicating that ratio images primarily contribute to more accurate lesion delineation once lesions have been successfully localized.

\subsection{Impact of MRI Representation Design}
Several important observations emerge from the experimental comparison.
First, FLAIR consistently outperformed T1w across all evaluation metrics, demonstrating that lesion-related signal abnormalities are more informative than anatomical features for automated FCD segmentation.
Second, the T1w/FLAIR ratio alone produced the lowest overall segmentation performance despite exhibiting the lowest false-positive rate among healthy controls. This suggests that ratio-derived representations should not replace conventional MRI sequences but rather complement them.
Third, integrating ratio-derived representations with conventional multimodal MRI consistently improved lesion delineation performance. In particular, E6 and E8 achieved the highest Dice scores among successfully detected lesions, indicating that ratio images provide complementary contrast information beyond standard T1w and FLAIR inputs.
These findings indicate that MRI representation design has a measurable influence on segmentation performance even when the underlying deep learning architecture remains unchanged.

\subsection{Representation Contribution Analysis}
To investigate the contribution of ratio-derived MRI representations, the multimodal experiments (E6--E8) were compared with the conventional T1w+FLAIR baseline (E5). This analysis isolates the effect of image representation while keeping the segmentation architecture, preprocessing pipeline, training strategy, and cross-validation protocol unchanged.

Adding the T1w/FLAIR ratio (E6) increased the overall Dice score from 0.358 to 0.367, while incorporating the FLAIR/T1w ratio (E7) produced a comparable Dice score of 0.361. Combining both ratio-derived representations (E8) achieved the highest overall Dice score (0.376), representing an absolute improvement of 0.018 over the conventional multimodal baseline. Similar trends were observed for sensitivity, which increased from 0.341 (E5) to 0.356 (E8), whereas precision remained comparable across all multimodal configurations.

A more pronounced improvement was observed when considering only successfully detected lesions. Compared with E5 (Dice = 0.534), E6 and E8 achieved Dice scores of 0.567 and 0.571, respectively, indicating that ratio-derived representations primarily enhance lesion delineation once lesions have been successfully localized. In contrast, lesion miss rates remained similar across the multimodal experiments (33--35\%), suggesting that the principal benefit of ratio-derived representations lies in improving segmentation quality rather than increasing lesion detection sensitivity.


\subsection{Qualitative Results}
\begin{figure*}[t]
    \centering
    \includegraphics[scale=0.58]{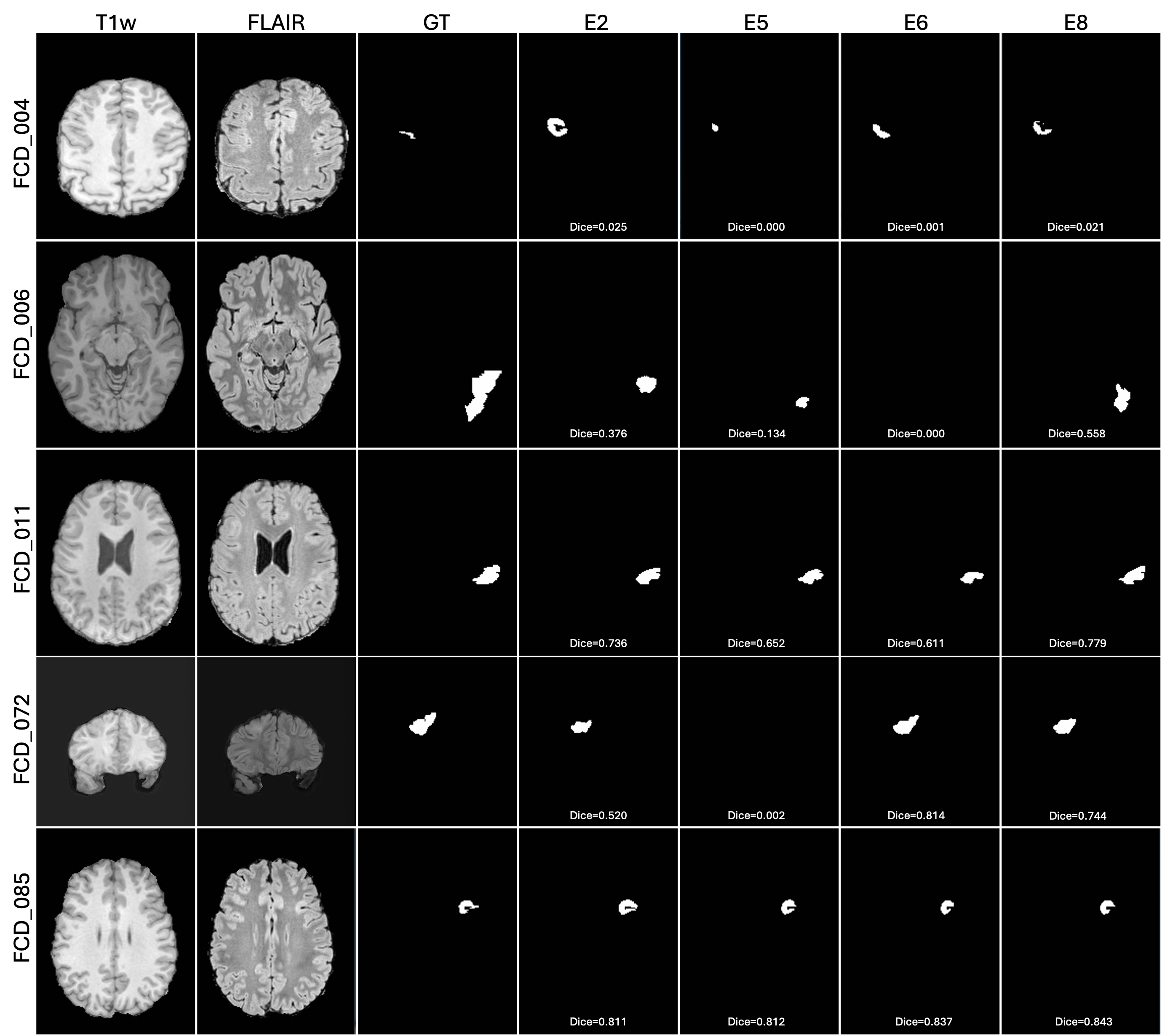}
    \caption{Representative qualitative segmentation results for five FCD subjects. For each subject (rows), the columns show the T1w image, FLAIR image, expert-annotated ground-truth (GT) lesion mask, and the predicted segmentations from E2 (FLAIR), E5 (T1w+FLAIR), E6 (E5+T1w/FLAIR), and E8 (E5+T1w/FLAIR+FLAIR/T1w). Dice scores are reported below each prediction.}
    \label{fig:qualitative}
\end{figure*}

Figure~\ref{fig:qualitative} presents representative qualitative segmentation results for five FCD subjects exhibiting varying lesion characteristics and segmentation difficulty. The selected examples include subtle lesions, challenging cases, and well-defined lesions, providing visual evidence of the influence of MRI representation on segmentation performance.

Consistent with the quantitative analysis, FLAIR (E2) generally achieved more accurate lesion localization than T1w alone, while the conventional multimodal representation (E5) further improved segmentation in several cases. Incorporating ratio-derived representations (E6 and E8) frequently resulted in more accurate lesion delineation and improved boundary agreement, particularly for subtle lesions. For example, the fourth case demonstrates a substantial improvement obtained by E6 over the conventional multimodal baseline, whereas the final case illustrates consistently high segmentation performance across all multimodal representations. Conversely, the first case remains challenging for all methods, highlighting the difficulty of detecting subtle FCD lesions and the inherent complexity of automated FCD segmentation.

\subsection{Representative Case Analysis}
Table~\ref{tab:case_examples} summarizes the Dice scores of five representative FCD subjects selected to illustrate different segmentation behaviours observed across the evaluated MRI representations. These cases include subtle lesions, challenging lesions, multimodal improvements, and well-segmented lesions, providing a subject-level perspective that complements the cohort-level analysis.
\begin{table*}[t]
\centering
\footnotesize
\setlength{\tabcolsep}{4pt}
\renewcommand{\arraystretch}{1}
\caption{Representative FCD cases demonstrating substantial performance differences across input configurations. The highest Dice score for each case is shown in bold.}
\label{tab:case_examples}
\begin{tabular}{l|cccccccc}
\toprule
Case & E1 & E2 & E3 & E4 & E5 & E6 & E7 & E8 \\
\midrule
FCD\_004 & 0.000 & 0.025 & 0.001 & 0.000 & 0.000 & 0.001 & 0.000 & \textbf{0.021} \\
FCD\_006 & 0.396 & 0.376 & 0.000 & \textbf{0.822} & 0.134 & 0.000 & 0.000 & 0.558 \\
FCD\_011 & 0.068 & 0.736 & 0.000 & 0.000 & 0.652 & 0.611 & 0.720 & \textbf{0.779} \\
FCD\_072 & 0.000 & 0.520 & 0.000 & 0.000 & 0.002 & \textbf{0.814} & 0.000 & 0.744 \\
FCD\_085 & 0.756 & 0.811 & 0.604 & 0.698 & 0.812 & 0.837 & 0.819 & \textbf{0.843} \\
\bottomrule
\end{tabular}
\end{table*}

Substantial variability in segmentation performance was observed across MRI representations. For example, FCD\_072 demonstrates one of the largest improvements obtained from ratio-derived representations, where the conventional multimodal baseline (E5) achieved a Dice score of only 0.002, whereas E6 and E8 achieved Dice scores of 0.814 and 0.744, respectively. Similarly, FCD\_011 shows consistent improvement with multimodal representations, with E8 achieving the highest Dice score (0.779).

Conversely, FCD\_004 remained challenging for all representations, highlighting the difficulty of segmenting subtle FCD lesions. FCD\_006 illustrates that different MRI representations emphasize distinct lesion characteristics, with FLAIR/T1w (E4) substantially outperforming the remaining configurations. Finally, FCD\_085 represents a relatively straightforward case in which all multimodal representations achieved high segmentation accuracy, with E8 producing the highest Dice score (0.843).


\subsection{Statistical Analysis}
A Friedman test revealed a significant overall effect of MRI representation on segmentation performance ($\chi^2$ = 74.16, p < 0.001). Post-hoc Wilcoxon signed-rank tests with FDR correction demonstrated that FLAIR significantly outperformed T1w (adjusted p = 0.0013), while FLAIR/T1w significantly outperformed T1w/FLAIR (adjusted p < 0.001). In contrast, although the multimodal ratio-enhanced representations (E6–E8) consistently achieved higher mean Dice scores than the conventional T1w+FLAIR baseline (E5), these improvements were not statistically significant after multiple-comparison correction (adjusted p > 0.05).



\section{Discussion}
This study systematically investigated the influence of MRI representations on deep learning-based FCD segmentation using a standardized nnU-Net framework. By maintaining identical preprocessing, network architecture, optimization strategy, and cross-validation protocol across all experiments, the observed performance differences can be attributed directly to the input MRI representations rather than architectural variations. A key finding of this study is that MRI representation design should be regarded as an integral component of deep learning model development rather than merely a preprocessing step. By fixing the network architecture throughout all experiments, we demonstrate that meaningful performance improvements can be achieved solely through representation design. This finding challenges the common assumption that improvements in FCD segmentation must primarily arise from architectural innovation.

\subsection{Impact of MRI Representations on FCD Segmentation}
Among the individual MRI representations, FLAIR consistently outperformed T1w and T1w/FLAIR, achieving the highest overall segmentation performance and the lowest lesion miss rate. This observation is consistent with the imaging characteristics of FCD, where cortical and subcortical signal abnormalities are often more conspicuous on FLAIR than on anatomical T1w images. In contrast, T1w primarily depicts cortical morphology and gray-white matter architecture, making subtle dysplastic lesions more difficult to distinguish from surrounding normal tissue. Although the FLAIR/T1w ratio improved lesion delineation for successfully detected lesions, neither ratio-derived representation alone surpassed conventional FLAIR, indicating that ratio images alone do not contain sufficient information for robust lesion localization.

The superior performance of FLAIR/T1w compared with T1w/FLAIR suggests that the ordering of the ratio operation influences the information presented to the segmentation network. FCD lesions are predominantly characterized by FLAIR hyperintensity, whereas T1w primarily provides anatomical context through cortical thickening and blurring of the gray-white matter junction. Using FLAIR as the numerator preserves lesion-related signal abnormalities while normalizing anatomical intensity variations, thereby enhancing lesion conspicuity. In contrast, placing FLAIR in the denominator attenuates lesion-related hyperintensity and produces a representation that emphasizes relative tissue contrast rather than pathological signal. This interpretation is supported by our results, in which FLAIR/T1w significantly outperformed T1w/FLAIR in both overall segmentation performance and lesion detection.

\subsection{Complementary Role of Ratio-Derived Representations}
Our experiments indicate that ratio-derived representations are most effective when used as complementary information rather than replacement modalities. This observation suggests that representation design should focus on enriching conventional MRI rather than substituting it.
The multimodal configurations consistently achieved higher Dice scores than the conventional T1w+FLAIR baseline, with the four-channel representation (E8) producing the best overall performance. The improvements were more evident when only successfully detected lesions were considered, suggesting that ratio-derived representations primarily refine lesion boundaries after localization rather than substantially improving lesion detection. This complementary behaviour indicates that ratio images enhance tissue contrast while conventional T1w and FLAIR preserve the anatomical and pathological information required for reliable lesion localization.

\subsection{Clinical Implications}
Accurate delineation of FCD lesions is essential for presurgical evaluation, where reliable estimation of lesion extent can support surgical planning and improve confidence in lesion localization. Although incorporating ratio-derived MRI representations produced modest average improvements that did not reach statistical significance after multiple-comparison correction, these gains were achieved without modifying the segmentation architecture or training strategy, highlighting MRI representation optimization as a simple and computationally efficient approach for improving automated FCD segmentation. Importantly, the subject-level analysis demonstrated that ratio-derived representations can provide substantial benefits for selected challenging cases. For example, the Dice score for FCD\_072 increased from 0.002 using the conventional multimodal representation (E5) to 0.814 with E6, indicating that representation design can markedly improve delineation for certain subtle lesions. These findings suggest that while ratio-derived representations do not uniformly improve every case, they have the potential to enhance segmentation of difficult FCD lesions encountered in clinical practice.

\subsection{Limitations and Future Work}

This study has certain limitations. The experiments were conducted using a single publicly available FCD dataset, and a standardized five-fold cross-validation protocol was employed. Nevertheless, validation on larger and more diverse multi-centre datasets is required to assess the generalizability of the findings across different scanners, acquisition protocols, and patient populations. Furthermore, this work intentionally focused on MRI representation analysis while using a fixed segmentation framework. Future studies should investigate learning MRI representations directly from multimodal data rather than relying on manually designed ratio-derived images.


\section{Conclusion}
This study demonstrates that MRI representation design is an important but largely overlooked component of deep learning-based FCD segmentation. Through a controlled benchmark using identical nnU-Net architectures, preprocessing, optimization strategy, and cross-validation protocols, we show that optimizing MRI representations is also a promising strategies for improving lesion delineation without increasing model complexity or altering the underlying segmentation framework. These findings suggest that future research should jointly optimize MRI representation design and network architecture rather than focusing exclusively on architectural innovation. Such a complementary approach may lead to more robust and clinically applicable segmentation systems while maintaining the simplicity and computational efficiency of established deep learning frameworks.


\section*{Code Availability}
The source code, preprocessing scripts, and experimental configurations used in this study are publicly available at:
\url{https://github.com/soumenca/FCD_Contrast_Analysis}

\bibliographystyle{IEEEtran}
\bibliography{references}

\end{document}